\documentclass[letterpaper, 10 pt, conference]{ieeeconf}  

\IEEEoverridecommandlockouts                              

\usepackage{graphics} 
\usepackage{epsfig} 
\usepackage{times} 
\usepackage{amsmath} 
\usepackage{amssymb}  
\usepackage{color}
\usepackage{booktabs}
\usepackage{multirow}
\usepackage{algorithm}
\usepackage{algorithmicx}
\usepackage{algpseudocode}
\usepackage{bm}
\usepackage{xcolor}
\usepackage{booktabs}
\usepackage{makecell}

\usepackage{balance}

\usepackage{marvosym}
\usepackage{ifsym}

\usepackage[
	colorlinks,
	anchorcolor=blue,
	linkcolor=cyan,
	urlcolor=red,
	citecolor=green,
	backref=page]{hyperref}
\usepackage[T1]{fontenc}

\title{\LARGE \bf 
CoopDETR: A Unified Cooperative Perception Framework for 3D Detection via Object Query
}

\author{Zhe Wang\textsuperscript{1}, Shaocong Xu\textsuperscript{1}, Xucai Zhuang\textsuperscript{1}, Tongda Xu\textsuperscript{1}, \\ Yan Wang\textsuperscript{1*}, Jingjing Liu\textsuperscript{1}, Yilun Chen\textsuperscript{1}, Ya-Qin Zhang\textsuperscript{1}\\
\thanks{$^{1}$Zhe Wang, Shaocong Xu, Xucai Zhuang, Tongda Xu, Yan Wang$^{*}$, Jingjing Liu, Yilun Chen, and Ya-Qin Zhang are with Institute for AI Industry Research (AIR), Tsinghua University, Beijing, China.
{\tt\small \{wangzhe wangyan\}@air.tsinghua.edu.cn}}
}

\begin{document}

\maketitle
\pagestyle{empty}
\thispagestyle{empty}


\begin{abstract}

Cooperative perception enhances the individual perception capabilities of autonomous vehicles (AVs) by providing a comprehensive view of the environment. However, balancing perception performance and transmission costs remains a significant challenge. Current approaches that transmit region-level features across agents are limited in interpretability and demand substantial bandwidth, making them unsuitable for practical applications. In this work, we propose CoopDETR, a novel cooperative perception framework that introduces object-level feature cooperation via object query. Our framework consists of two key modules: single-agent query generation, which efficiently encodes raw sensor data into object queries, reducing transmission cost while preserving essential information for detection; and cross-agent query fusion, which includes Spatial Query Matching (SQM) and Object Query Aggregation (OQA) to enable effective interaction between queries. Our experiments on the OPV2V and V2XSet datasets demonstrate that CoopDETR achieves state-of-the-art performance and significantly reduces transmission costs to 1/782 of previous methods.



\end{abstract}

\section{INTRODUCTION}

In recent years, autonomous driving has made significant progress, however, substantial challenges persist, particularly in the area of single-vehicle perception. Limitations in range and accuracy still affect the safety of autonomous vehicles. Cooperative perception, which allows vehicles and infrastructure to communicate, offers a potential solution. Cooperative perception addresses key limitations of single vehicles by providing more comprehensive and reliable information, particularly in complex traffic scenarios. As a result, this approach has garnered increasing attention from researchers recently.~\cite{han2023collaborative, liu2023towardsv2x_survey}.


\begin{figure}[t]
	\centering  
	\includegraphics[width=0.9\linewidth]{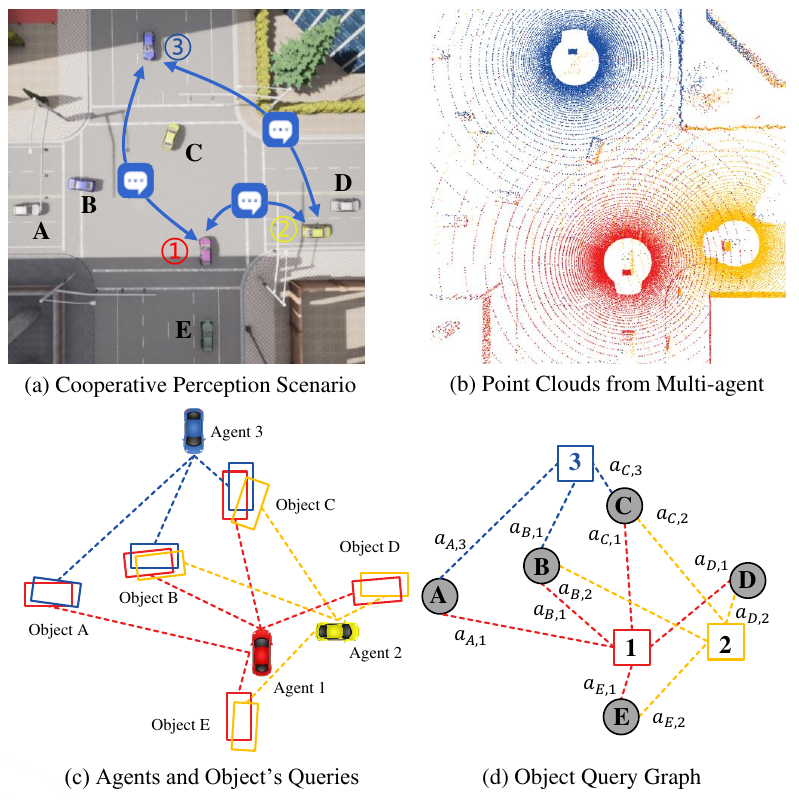} 
	\caption{Consider a typical cooperative perception scenario involving three connected agents (1 to 3) and five objects (A to E) to be detected. Each agent processes its respective point cloud and generates queries of surrounding objects using a DETR-based model. Queries corresponding to the same object in the scene can be connected to form an object query graph, facilitating further query fusion via attention mechanism. Subfigure (d) illustrates the object query graphs for objects A to E.}  
	\label{fig:Teaser}   
\end{figure}

Compared to traditional single-vehicle perception, cooperative perception introduces a set of new challenges. The foremost issue lies in determining what information should be transmitted to keep a balance between perception performance and transmission cost. Based on the type of data communicated among agents, current cooperative perception methods can be categorized into three typical cooperation paradigms: \textit{early fusion} (EF) of raw sensor data~\cite{yu2022dairv2x,chen2022co3,chen2019cooper}, \textit{intermediate fusion} (IF) of features~\cite{mehr2019disconet,xu2022opv2v,wang2020v2vnet,fan2023quest,hu2022where2comm,xu2022v2xvit}, and \textit{late fusion} (LF) of prediction results~\cite{yu2022dairv2x,chen2022model-agnostic}.

Early fusion involves the transmission of raw sensor data from each agent, which preserves the original information but requires significantly higher transmission costs for cooperation. In contrast, late fusion has the advantage of reduced transmission costs, making it suitable for practical applications. However, it suffers from considerable information loss from the source sensor data, and its performance is highly contingent on the perception accuracy of individual agents. Intermediate fusion keeps a balance between performance and bandwidth, as features can be compressed for lower transmission costs while retaining critical information extracted from raw data. Several approaches~\cite{hu2022where2comm,xu2022v2xvit} have encoded raw data into dense, region-level features for communication, such as Bird's-eye-view (BEV) features. However, this type of representation, which aims to depict the entire scene, suffers from limited interpretability and may still contain redundant information despite feature selection mechanisms~\cite{hu2022where2comm}. Considering that the most valuable information for 3D detection is object-specific, can we design a communication mechanism that is centered on object-level features specifically tailored for cooperative perception?

Inspired by transformer-based object detection methods (DETR)~\cite{carion2020detr,wang2022detr3d,chen2022futr3d,fan2023quest}, we propose a novel paradigm for achieving object-level feature cooperation in multi-agent systems, where raw data is encoded into queries, with each query corresponding to a specific object in the scene.
This approach offers the similar interpretability as late fusion while achieving lower communication bandwidth compared to early fusion and region-level intermediate fusion.
The proposed framework, named CoopDETR, introduces a unified cooperative perception model that leverages object queries to facilitate interaction across multiple agents.
As illustrated in Figure~\ref{fig:Teaser}, in a simplified scene, each agent generates a set of queries for objects within its observation range so that different agents produce distinct queries for the same object, reflecting diverse information.
An object query graph can be constructed for each object, where each node represents a query derived from a single agent's sensor data. These queries can be flexibly fused to aggregate information, enabling a more comprehensive representation of the object. Each query can then be decoded to infer the object's category and bounding box.

Specifically, CoopDETR consists of two primary modules: single-agent query generation and cross-agent query fusion. In the query generation module, each agent leverages a transformer-based model, PointDETR, to update queries based on point cloud features, which can then be shared with other agents. In the cross-agent query fusion module, upon receiving queries from other agents, the ego agent applies Spatial Query Matching (SQM) to associate similar queries and cluster them into distinct object query graphs. These queries within the same graph are subsequently fused through an attention-based mechanism in the Object Query Aggregation (OQA) process. The fused queries are subsequently fed into detection heads for final prediction. Experimental results on the OPV2V~\cite{xu2022opv2v} and V2XSet~\cite{xu2022v2xvit} datasets demonstrate that CoopDETR achieves an improved trade-off between perception accuracy and transmission cost.

\begin{itemize}
    \item We propose CoopDETR, a novel cooperative perception framework based on object query, achieving more efficient communication through object-level feature cooperation, compared to dense BEV features for scene-level feature cooperation.
    \item We design Spatial Query Matching (SQM) and Object Query Aggregation (OQA) modules for query interaction to select queries of co-aware objects and fuse queries at the instance level.
    \item We achieve state-of-the-art results on the V2XSet and OPV2V datasets, outperforming other cooperative perception methods while reducing transmission costs to 1/782 of previous methods.
\end{itemize}

\section{Related Work}
\subsection{V2X Cooperative Perception}

Current research on cooperative perception mainly aims to extend the perception range and improve perception capability of autonomous vehicles~\cite {wang2024emiff,han2023collaborative}. The most intuitive approach is Early Fusion, which transmits raw sensor data~\cite{chen2019cooper,chen2022co3}. However, transmitting raw data requires high transmission costs, making it impractical for real-world deployment. Late Fusion that transmits perception results from each agent is the most bandwidth-efficient paradigm~\cite{chen2022model-agnostic,yu2022dairv2x}. Yet, its performance relies heavily on the accuracy of each agent's perception result. Most research has shifted toward intermediate fusion, which transmits region-level features for better performance-bandwidth balance~\cite{xu2022opv2v,mehr2019disconet,xu2022v2xvit,hu2022where2comm,xu2022cobevt,wang2020v2vnet,yu2023ffnet,cui2022coopernaut,hu2023coca3d,wang2024emiff,wei2023cobevflow}. Although these methods incorporate strategies to reduce transmission costs, such as feature selection via spatial confidence maps~\cite{hu2022where2comm}, feature compression~\cite{xu2022opv2v,xu2022v2xvit,wang2024emiff}, and flow-based prediction~\cite{yu2023ffnet,wei2023cobevflow}, region-level feature is still redundant for object detection and lacks interpretability~\cite{fan2023quest}. QUEST \cite{fan2023quest} proposes the concept of query-cooperation paradigm but focuses only on a simple V2I scenario involving one vehicle and infrastructure.  To enable more efficient cooperation across multi-agent systems, we propose a unified cooperation perception framework that transmits object-level queries across agents.

\begin{figure*}[t]
	\centering  
	\includegraphics[width=0.9\linewidth]{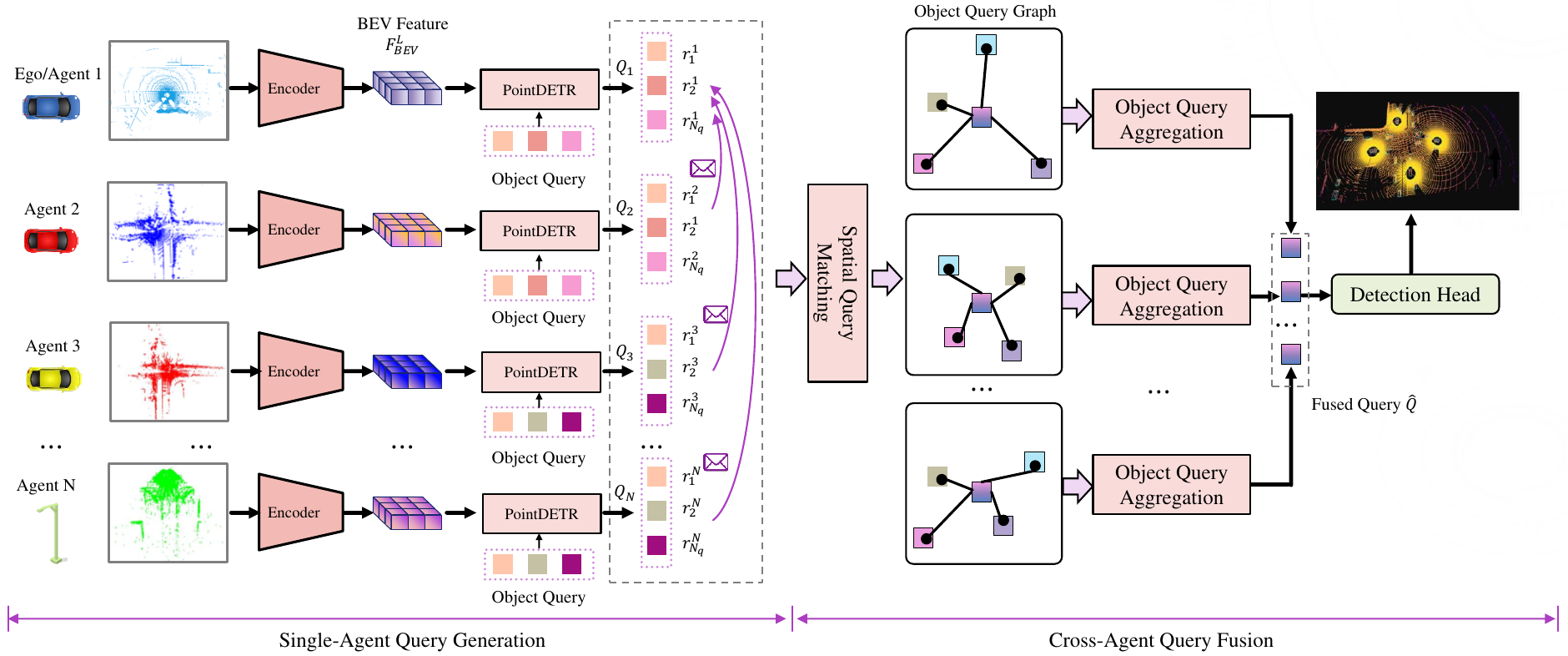} 
	\caption{The general framework of CoopDETR. For each agent, the query generation module learns $N_q$ object queries from raw data. Each object in the scene will correspond to a query. For the whole multi-agent system, one object may be observed by different agents and be associated with different queries. Take $i$-th agent as ego agent, object queries $Q_{j} = \{q^{j}_{1},\dots,q^{j}_{N_q}\}$ from $j$-th agent and their reference points $r$ will be transmitted to $i$-th agent. In cross-agent query fusion module, all queries will be fused with two steps, the 
 the first step is to associate different queries for co-aware objects through spatial query matching (SQM) and generate object query graph for each object. The second step is to fuse all queries in the same graph using Object Query aggregation (OQA) and generate a set of updated queries $\hat{Q}$, which will be fed to detection heads for category and bounding box prediction. }  
	\label{fig:framework}   
\end{figure*}

\subsection{Transformer-based Perception}
The pioneering work DETR~\cite{carion2020detr} regards 2D object detection task as a set-to-set problem. The query mechanism has been increasingly adopted across various perception tasks, including 3D object detection~\cite{wang2022detr3d,chen2022futr3d,chen2023transiff,fan2023quest}, object tracking~\cite{zeng2022motr,zhang2022mutr3d,pang2023pf-track,meinhardt2022trackformer}, semantic segmentatin~\cite{li2022bevformer,peng2022bevsegformer,maiti2023transfusion}, and planning~\cite{hu2023_uniad,yu2024_univ2x}. Query-based approaches typically leverage sparse, learnable queries for attentive feature aggregation to capture complex relationships among sequential elements. FUTR3D~\cite{chen2022futr3d} predicts 3D query locations and retrieves corresponding multi-modal features from cameras, LiDARs, and radars via projection. BEVFormer~\cite{li2022bevformer,yang2023bevformerv2} introduces grid-shaped queries in BEV and updates them by interacting with spatio-temporal features using deformable transformers. While most existing query-based methods focus on individual perception, QUEST~\cite{fan2023quest} and TransIFF~\cite{chen2023transiff} extend it to vehicle-to-infrastructure (V2I) scenarios. In this work, we introduce a novel query fusion mechanism, which facilitates efficient query matching and aggregation tailored for multi-agent systems.

\section{Method}

 Considering $N$ agents in a multi-agent system, $\boldsymbol{\mathcal{X}}_{i}$ is the point clouds observed by $i$-th agent and $\boldsymbol{\mathcal{Y}}_{i}$ is the corresponding perception supervision. The object of CoopDETR is to achieve the maximized perception performance of all agents with a communication budget of $B$.

\begin{equation}
    \begin{split}
    \xi(B)  = \underset{\theta, \boldsymbol{M}_{j\rightarrow i}} {\arg \max } & \sum_i^N g\left(\Psi_\theta\left(\boldsymbol{\mathcal{X}}_{i},\left\{{\boldsymbol{M}}_{j\rightarrow i}\right\}_{j=1}^N\right), \boldsymbol{\mathcal{Y}_{i}}\right) \\
     \text { s.t. } & \sum_j \left| {\boldsymbol{M}}_{j \rightarrow i}\right| \leq B
    \end{split}
\end{equation}
where $g(\cdot,\cdot)$ denotes the perception evaluation metric and $\Psi_\theta$ is the perception network with trainable parameter $\theta$, and ${\boldsymbol{M}}_{j\rightarrow i}$ means the message transmitted from the $j$-th agent to the $i$-th agent. 

The architecture of CoopDETR is shown in Figure~\ref{fig:framework}, which includes single-agent query generation and cross-agent query fusion modules.

\begin{figure}[ht]
	\centering  
        \includegraphics[width=0.9\linewidth]{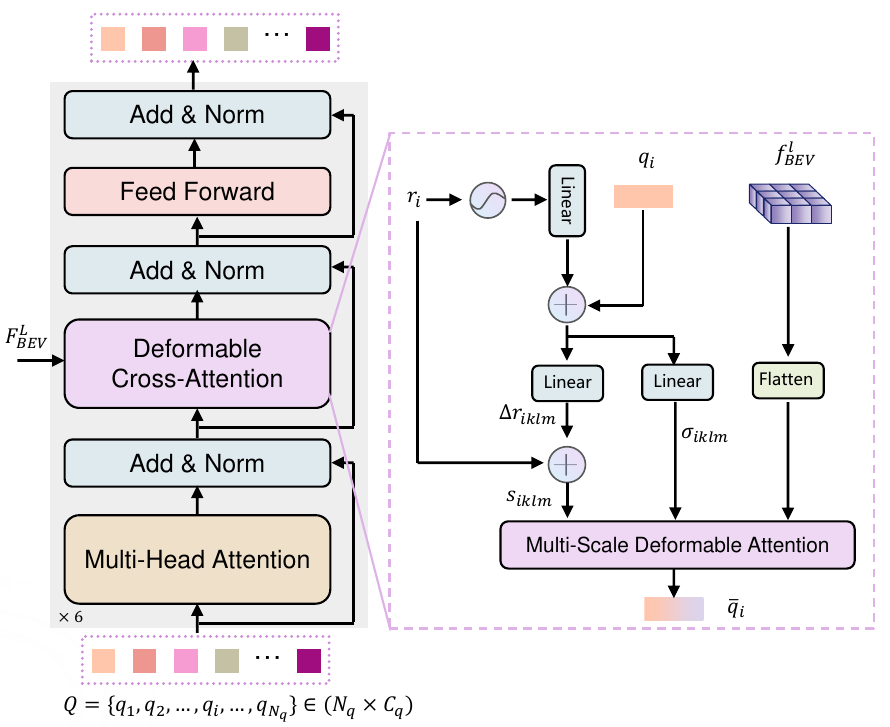} 
	\caption{Illustration of PointDETR module.
 }
	\label{fig:PointDETR}   
\end{figure}

\subsection{Single-Agent Query Generation}

\textbf{Feature Encoder} Following FUTR3D~\cite{chen2022futr3d}, we employ PointPillar~\cite{lang2019pointpillars} as the feature encoder to process the point cloud. After passing through the 3D backbone and FPN~\cite{lin2017fpn}, we extract multi-scale bird’s-eye view (BEV) features, denoted as \( \textbf{F}^{L}_{BEV} = [F^{l}_{BEV} \in \mathbb{R}^{C \times H_{l} \times W_{l}}]^{L}_{l=1} \), where \( L \) denotes the number of feature levels.

\textbf{PointDETR} Each agent initializes a set of object queries $Q \in \mathbb{R}^{N_q \times C_q} $, which are then refined dynamically using the transformer decoder, PointDETR (illustrated in Figure~\ref{fig:PointDETR}). A key aspect of PointDETR lies in determining how to sample features corresponding to an object query from BEV features.

In the Deformable Cross-Attention block, a reference point \( r_{i} \in \mathbb{R}^{3} \), representing the center of the \(i\)-th object's bounding box, is decoded from an object query using a linear neural network. This reference point is initialized from a sketch and is used to sample BEV features through multi-scale deformable attention~\cite {zhu2020deformable}. As illustrated in Figure~\ref{fig:PointDETR}, the reference point \( r_{i} \) is encoded into a positional embedding using a sine function and a linear network, which are subsequently added to the query \( q_{i} \). The sample location offset \( \Delta r_{iklm} \) and attention weight \( \sigma_{i k l m} \) are generated through separate linear networks. The sample offset \( \Delta r_{iklm} \) is added to the initial reference point \( r_{i} \) to obtain the final sample location \( s_{iklm} \). The BEV features are then sampled at these locations \( s_{iklm} \) using bilinear sampling. Sample location $s_{iklm} \in [0,1]^2$ is represented by normalized coordinates of feature map.

\begin{equation}
\boldsymbol{V}_{i k l m} = f^{bilinear}(f^{l}_{BEV},s_{iklm})
\end{equation}

Where $k$ indexes the sampling point, $l$ indexes the lidar feature's scale, and $m$ indexes the attention head. Following Deformable DETR~\cite{zhu2020deformable},  we samples $L \times K$ points from multi-scale features and update queries by

\begin{equation}
\Bar{q_{i}}=\sum_{m=1}^M \boldsymbol{W}_{i}\left[\sum_{l=1}^L \sum_{k=1}^K \sigma_{i k l m} \cdot \boldsymbol{W}_{i}^{\prime} \boldsymbol{V}_{i k l m}\right]
\end{equation}

The scalar attention weight $\sigma_{iklm}$ is normalized by $\sum_{l=1}^L \sum_{k=1}^K \sigma_{i k l m}=1$.

\subsection{Cross-Agent Query Fusion}

\begin{figure}[ht]
	\centering  
        \includegraphics[width=0.9\linewidth]{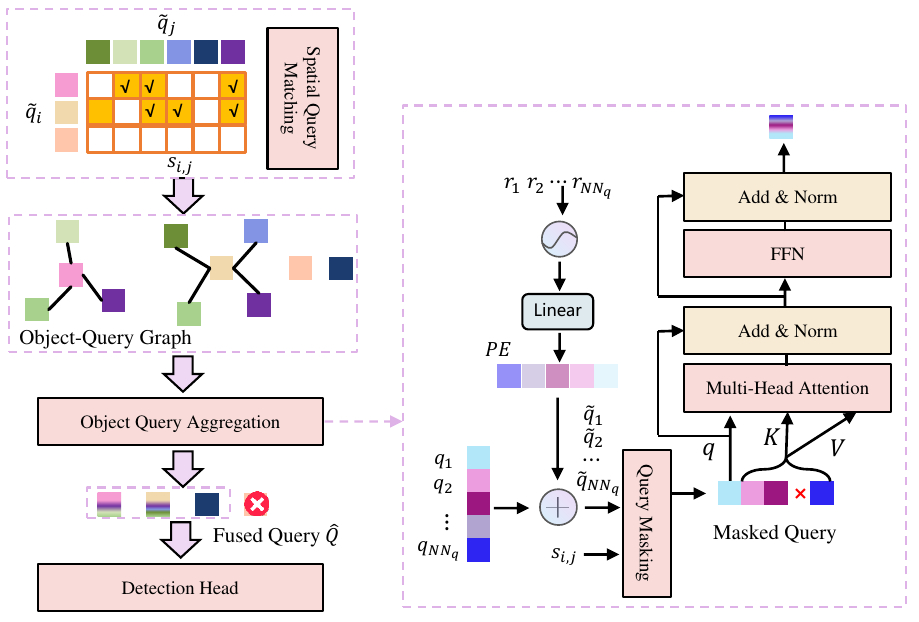} 
	\caption{The details of Cross-Agent Query Fusion.
 }
	\label{fig:QF}   
\end{figure}

After communication, ego agent $i$ has $NN_{q}$ queries. All objects encoded by these queries can be categorized into three types: those co-aware to both ego agent and other agents, those observed solely by ego agent, and those recognized only by other agents. As shown in Figure~\ref{fig:QF}, Spatial Query Matching can associate queries corresponding to the same object and generate an object query graph via masked attention. Object Query Aggregation fuses all queries in the same graph. All fused queries will be permuted by their own confidence score and only the maximum $N_{q}$ ones will be fed into the detection head for prediction. The detection head is identical to that used in FUTR3D~\cite{chen2022futr3d}, comprising two separate MLP for classification and bounding box prediction respectively. Following ~\cite{wang2022detr3d,chen2022futr3d,carion2020detr}, we compute a set-to-set loss between predictions and ground-truths.


\textbf{Spatial Query Matching} To associate queries related to the same object, it is necessary to measure the similarity between the $i$-th query from the ego agent and the $j$-th query from other agents. Directly using reference points for matching suffers
 from inevitable pose errors. Therefore, we combine the query feature $q_{i}$, which contains contextual information from point cloud, with the position embedding derived from the reference point $r_{i}$ for more accurate matching. The refined query, $\Tilde{q}_{i}$, is obtained as follows:


\begin{equation}
\Tilde{q}_{i} = q_{i} + \Phi(f_{sin}(r_{i}))
\end{equation}
where $\Phi$ means linear neural network. $f_{sin}$ means sine function in Transformer~\cite{vaswani2017attention}. $\Phi(f_{sin}(r_{i}))$ denote position embedding (PE) of $r_{i}$. The similarity of refined queries $\Tilde{q}_{i}$ and $\Tilde{q}_{j}$ is calculated by:

\begin{equation}
     s_{i,j} = \text{sigmoid} \left( \frac{\langle \Tilde{q}_{i},\Tilde{q}_{j} \rangle}{\Vert \Tilde{q}_{i} \Vert_{2} \cdot \Vert \Tilde{q}_{j} \Vert_{2}}  \right)
\end{equation}
$\langle \cdot \rangle$ means inner product. We set threshold $\mu$ to determine whether two queries should be associated together. 

\textbf{Object Query Aggregation} After Query After Query Matching, we obtain multiple object query graphs, each corresponding to an object in the scene.  Queries in the same graph are aggregated via multi-head attention~\cite{vaswani2017attention}. Take one graph as an example, query $\Tilde{q}_{j}$ with similarity $s_{i,j}$ below the threshold is masked for calculation of attention weights. Ego agent's query is encoded as $q\in \mathbb{R}^{1 \times C}$ and masked query from other agents are encoded as key $K$ and value $V$. The fused query $\Hat{q}_{i}$ can be calculated by

\begin{equation}
\Hat{q} = \operatorname{Mask\_Attention}(q, K, V)=\operatorname{softmax}\left(\frac{\epsilon (q K^T)}{\sqrt{d_k}}\right) V
\end{equation}
where $\epsilon$ denotes masking operation. The attention weight between the key $K$ for the masked queries and the query $q$ is set to zero, ensuring that the update of the ego agent's query does not involve the masked queries.

\section{Expermients}

\subsection{Implementation Details}

\textbf{Datasets.}
To evaluate the performance of CoopDETR, we conduct extensive experiments on two benchmark datasets for the multi-agent cooperative perception task: OPV2V~\cite{xu2022opv2v} and V2XSet~\cite{xu2022v2xvit}. OPV2V is a large-scale simulated dataset collected in vehicle-to-vehicle (V2V) scenarios, consisting of 11,464 frames of point clouds with corresponding 3D annotations. The dataset is split into 6,764 training frames, 1,981 validation frames, and 2,719 testing frames. V2XSet~\cite{xu2022v2xvit}, co-simulated by CARLA~\cite{dosovitskiy2017carla} and OpenCDA~\cite{xu2021opencda}, is designed for V2X cooperative perception. It includes 73 representative scenes with 2 to 5 connected agents, totaling 11,447 frames of annotated LiDAR point clouds, with splits of 6,694 frames for training, 1,920 for validation, and 2,833 for testing.


\textbf{Evaluation Metrics.}  We use Average Precision (AP) at Intersection-over-Union (IoU) thresholds of 0.3, 0.5, and 0.7 to evaluate 3D object detection performance. For evaluating transmission cost, we follow the method outlined in ~\cite{hu2022where2comm}, which calculates communication volume by measuring the message size in Bytes, represented on a logarithmic scale with base 2.

\textbf{Experimental Settings.} 
The detection range of the LiDAR is set to $[-140.8\text{m}, 140.8\text{m}]$ along the X-axis, $[-40.0\text{m}, 40.0\text{m}]$ along the Y-axis, and $[-3.0\text{m}, 1.0\text{m}]$ along the Z-axis. Pillar size in PointPillar~\cite{lang2019pointpillars} is $[0.2\text{m}, 0.2\text{m}]$. The maximum feature map size from the feature encoder is $[H_{1}, W_{1}] = [400, 1408]$ with 256 channels. In PointDETR, the number of queries $N_{q}$ is 180, the number of sampled points $K$ is 4, and the number of attention heads $M$ is 8. The number of point cloud feature scales $L$ is set to 4, and the threshold $\mu$ for query matching is 0.3. We train the model for 50 epochs on both the OPV2V and V2XSet datasets, with a batch size of 4 for each. Training is conducted on NVIDIA Tesla A30 GPUs.
We employ AdamW~\cite{loshchilov2017adamw} as the optimizer, with a weight decay of $10^{-2}$. The learning rate is set to $2 \times 10^{-4}$, and we utilize a cosine annealing learning rate scheduler~\cite{loshchilov2016coslrs} with 10 warm-up epochs and a warm-up learning rate of $10^{-5}$.

\begin{table}[htbp]
    \centering
    \footnotesize
    \begin{tabular}{ccc}
    \toprule
    \multirow{2}{*}{\textbf{Model}} 
     & \textbf{V2XSet} & \textbf{OPV2V} \\ 
     \cmidrule(lr){2-3}
      & \textbf{AP@0.5/0.7} & \textbf{AP@0.5/0.7} \\
    \midrule
    No Fusion  & 60.60/40.20 & 68.71/48.66 \\
    Late Fusion  & 66.79/50.95 & 82.24/65.78 \\
    Early Fusion  & 77.39/50.45 & 81.30/68.69 \\
    When2com \cite{Liu2020when2com}  & 70.16/53.72 & 77.85/62.40 \\
    V2VNet \cite{wang2020v2vnet}  & 81.80/61.35 & 82.79/70.31 \\
    AttFuse \cite{xu2022opv2v}  & 76.27/57.93 & 83.21/70.09 \\
    V2X-ViT \cite{xu2022v2xvit}  & 85.13/68.67 & 86.72/74.94 \\
    DiscoNet \cite{mehr2019disconet} & 82.18/63.73 & 87.38/73.19 \\
    CoBEVT \cite{xu2022cobevt}  & 83.01/62.67 & 87.40/74.35 \\
    Where2comm \cite{hu2022where2comm}  & 85.78/72.42 & 88.07/75.06 \\
    \midrule
    \textbf{CoopDETR (Ours)}  & \textbf{86.96/76.51} & \textbf{90.59/83.97} \\
    \bottomrule
    \end{tabular}
    \caption{Performance comparison on the V2XSet, and OPV2V datasets. The results are reported in AP@0.5/0.7.}
    \label{TAB:ap50}
\end{table}

\subsection{Quantitative Evaluation}



\begin{figure}[htbp]
	\centering  
	\includegraphics[width=0.9\linewidth]{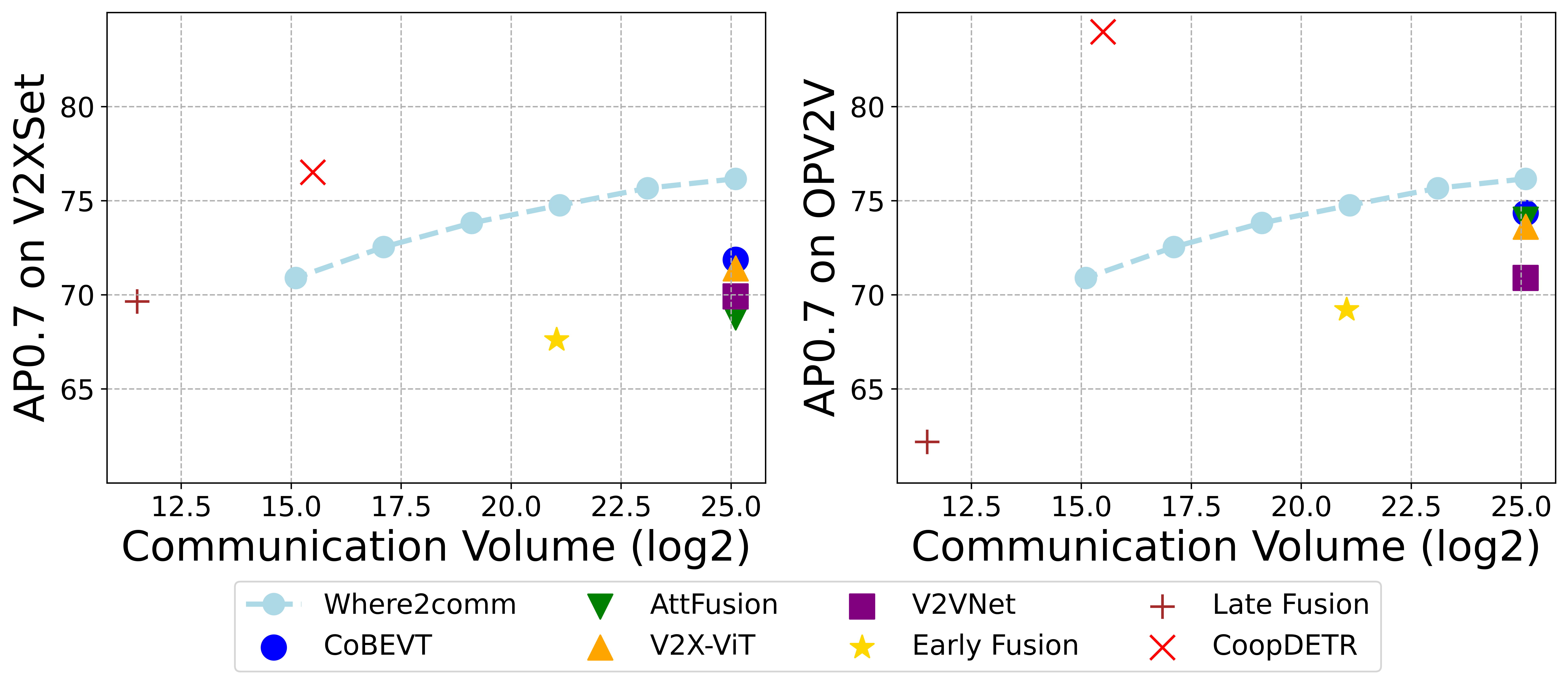} 
	\caption{Cooperative perception performance comparison of CoopDETR and other methods on V2XSet and OPV2V dataset. The communication volumes are also depicted in this figure.}  
	\label{fig:Com_Volume}   
\end{figure}

\textbf{Object Detection Results} We compare the detection performance of the proposed CoopDETR with various cooperative perception methods on the OPV2V and V2XSet datasets, as shown in Table~\ref{TAB:ap50}. CoopDETR is compared with state-of-the-art methods that focus on single-frame fusion, including When2com~\cite{Liu2020when2com}, V2VNet~\cite{wang2020v2vnet}, AttFuse~\cite{xu2022opv2v}, V2X-ViT~\cite{xu2022v2xvit}, DiscoNet~\cite{mehr2019disconet}, CoBEVT~\cite{xu2022cobevt}, and Where2comm~\cite{hu2022where2comm}. CoopDETR significantly outperforms these previous methods, demonstrating the superiority of our model. In particular, CoopDETR improves the state-of-the-art AP\@ 0.7 performance on the OPV2V and V2XSet datasets by 8.91 and 4.09, respectively, highlighting the effectiveness of the object-level fusion paradigm.

\textbf{Transmission Cost} The performance comparison results with distinct transmission costs, are shown in Figure~\ref{fig:Com_Volume}. The blue curves represent the detection performance of Where2comm at different compression rates, while the red cross marks CoopDETR, which achieves both lower communication volume and better performance than other intermediate fusion methods. Notably, CoopDETR's transmission volume is just 1/782 of that used by other intermediate fusion methods, and it approaches the volume of Late Fusion, typically regarded as the lower bound for transmission cost. These significant improvements underscore the effectiveness of CoopDETR's query-based mechanism in filtering redundant information from dense point cloud features, enabling it to learn more precise object representations.

\begin{figure}[htbp]
	\centering  
	\includegraphics[width=\linewidth]{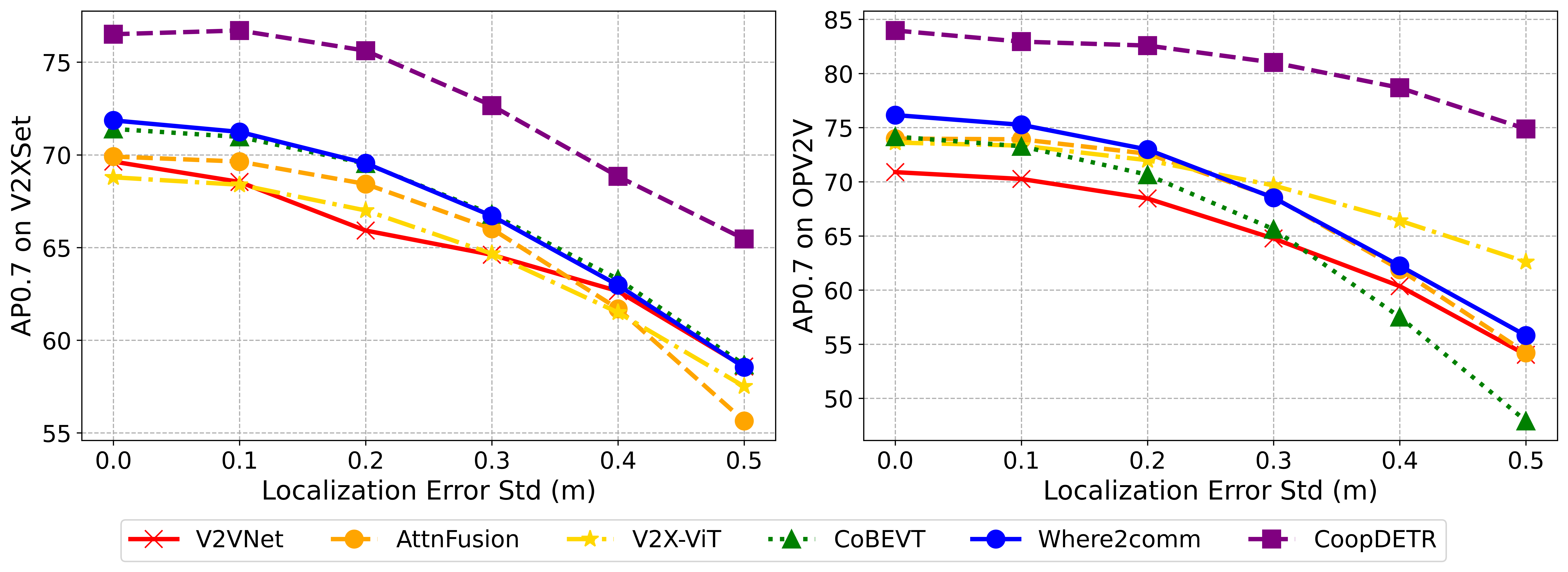} 
	\caption{Models' robustness to the localization error on the V2XSet and OPV2V datasets.}  
	\label{fig:Localization_Error}   
\end{figure}

\begin{figure}[htbp]
	\centering  
	\includegraphics[width=\linewidth]{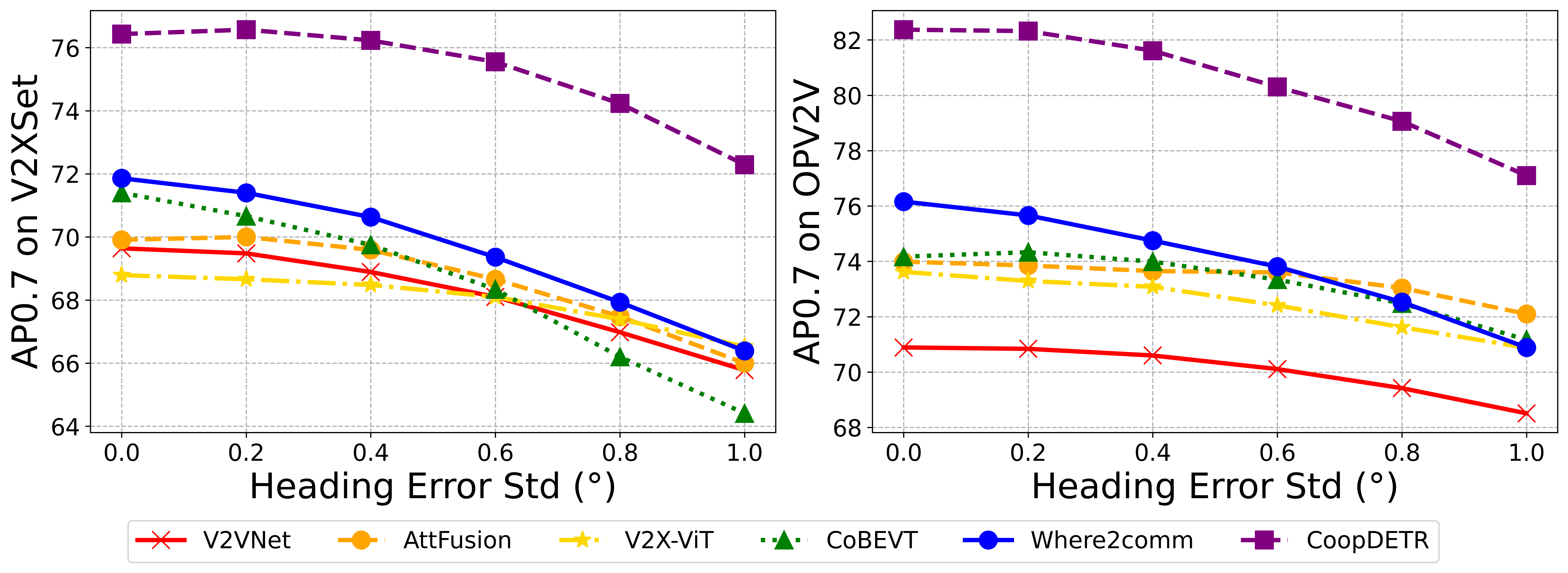} 
	\caption{Models' robustness to the heading error on the V2XSet and OPV2V datasets.}  
	\label{fig:Heading_Error}   
\end{figure}

\begin{figure*}[t]
	\centering  
	\includegraphics[width=0.85\linewidth]{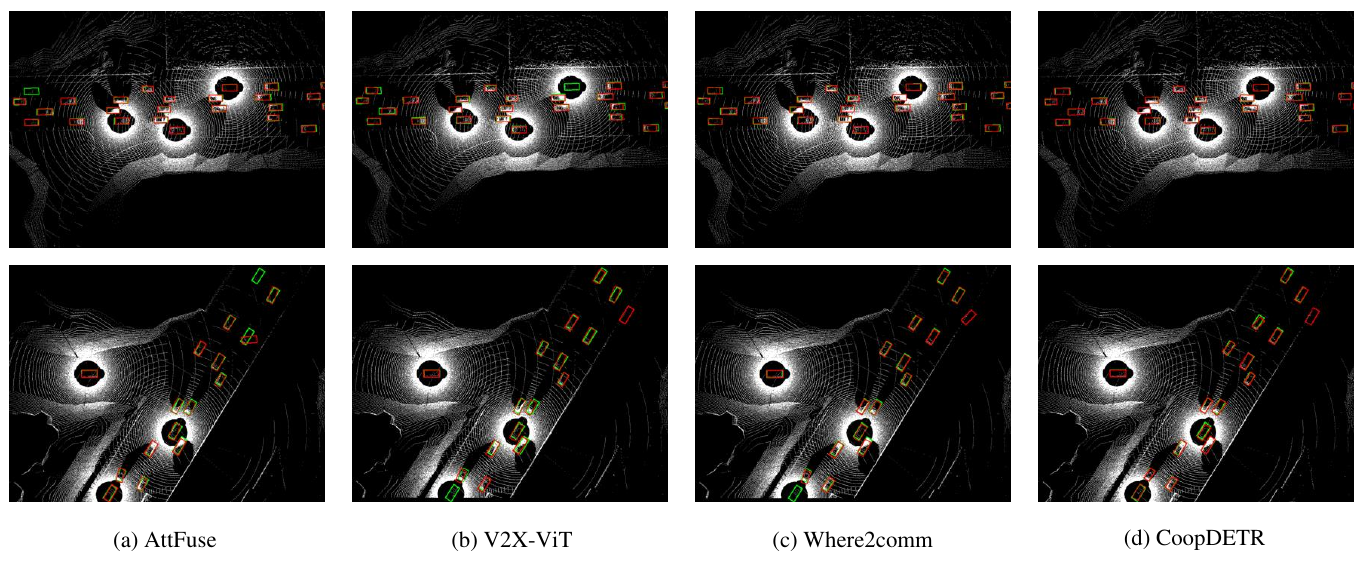} 
	\caption{Visualization results from the OPV2V dataset. Green and red bounding boxes denote the ground truths and prediction results respectively. CoopDETR qualitatively outperforms other cooperative perception methods AttFuse~\cite{xu2022opv2v}, V2X-ViT~\cite{xu2022v2xvit}, and Where2comm~\cite{hu2022where2comm}. Our method yields more accurate detection results and has fewer false positive objects than others.}
	\label{fig:vis_results}   
\end{figure*}

\textbf{Robustness to Localization and Heading Errors}. To evaluate the detection performance of CoopDETR under realistic scenarios, we follow ~\cite{yang2024how2comm} to simulate pose errors that occur during communication between agents. The results are shown in Figures~\ref{fig:Localization_Error} and~\ref{fig:Heading_Error}. Localization and heading errors, sampled from Gaussian distributions with standard deviations of $\sigma_{xyz} \in [0m, 0.5m]$ and $\sigma_{h} \in [0^{\circ}, 1.0^{\circ}]$, respectively, are introduced to the agents' locations. The figures reveal a consistent degradation in the performance of all cooperative perception methods, caused by increased feature misalignment. Notably, CoopDETR outperforms previous models across both datasets at all error levels, demonstrating its robustness against pose noise in communication processing. PointDETR, by employing deformable attention to integrate query and point cloud features, helps mitigate the negative effects of pose errors.


\subsection{Ablation Study}

Table~\ref{TAB:AB} shows ablation studies on all datasets to understand the necessity of model designs and strategies in CoopDETR. 

\textbf{Impact of Feature Size} The resolution of the pillar directly impacts the size of feature output from the point cloud encoder. Lower resolution leads to larger features and a more complex encoder, resulting in detailed representations and better performance. As seen in~\ref{TAB:AB}, AttFuse performs better with larger pillars, this also increases communication volume (CV). Regardless of feature size, a fixed number of queries can still interact with the features and receive updates, facilitating efficient feature communication. This object-level fusion paradigm enables agents to use encoders with different pillar sizes based on their computational capacity, and the queries can be fused in a unified framework.


\textbf{Impact of Query Number} Empirically, we experiment with varying the number of queries $N_q$ in PointDETR and found that 180 queries deliver the most competitive detection performance. An excessive number of queries can lead to performance bottlenecks by increasing the difficulty of model convergence and raising transmission costs. Too few queries may result in information loss during the interaction between queries and BEV features.


\textbf{Importance of SQM} We remove the Spatial Query Masking (SQM) in CoopDETR, where all queries are concatenated and fed directly into the detection head. In this setup, each query interacts with all other queries without masking during query aggregation. This increases the difficulty of model convergence and introduces redundant information into the fusion process. By contrast, SQM enables more efficient integration of contextual and spatial information within queries, improving both performance and convergence.

\begin{table}[ht]
  \footnotesize
  \centering
  \resizebox{\linewidth}{!}{
    \begin{tabular}{ccccc}
    \hline
        \multirow{2}{*}{Designs} & V2XSet & OPV2V & \multirow{2}{*}{$H_{1} \times W_{1}$ } & \multirow{2}{*}{CV} \\
                                 & AP@0.7 & AP@0.7 \\ \midrule
         \multicolumn{5}{c}{Impact of Feature Size} \\ 
         \midrule
        \multirow{3}{*}{CoopDETR}               &  69.94      & 76.06     & 200x704 & 0.046MB  \\
                        &  75.62      & 81.04     & 400x1408 & 0.046MB   \\ 
                       &  76.71      & 82.34     & 800x2816 & 0.046MB   \\ \hline
        \multirow{3}{*}{AttFuse}                &  68.92      & 70.09   & 200x704 & 36.044MB    \\ 
                        &  69.64      & 73.92   & 400x1408 & 144.179MB    \\
                        &  69.91      & 73.99   & 800x2816 & 576.717MB    \\
        \midrule
        \multicolumn{5}{c}{Impact of Query Number$N_{q}$} \\
        \midrule
        $N_{q}=90$                &  69.04      & 82.31     & \multirow{6}{*}{400x1408} & 0.023MB\\
        $N_{q}=180$ (Default)                &  \textbf{76.51}      & \textbf{83.97}      &  & 0.046MB\\
        $N_{q}=360$               &  69.51     & 80.07     & & 0.092MB \\ 
        $N_{q}=540$               &  71.72      & 78.15       & &0.138MB\\ 
        $N_{q}=720$                &  67.74      & 81.82       & & 0.184MB\\ 
        $N_{q}=900$                &  71.54      & 81.49       & & 0.230MB\\ 
        \midrule
        \multicolumn{5}{c}{Importance of SQM} \\
        \midrule
        w/o SQM               &  69.94      & 80.40  &\multirow{2}{*}{400x1408} &\multirow{2}{*}{0.046MB}    \\
        w SQM                &  \textbf{76.51}      & \textbf{83.97}   & &    \\ \hline
        
    \end{tabular}}
    \caption{Analysis on the choice of query number, pillar size, and component of query matching. "w/o" means without.}
    \label{TAB:AB}
\end{table}

\subsection{Quanlitative Evaluation} 
To illustrate the perception performance of various models, Figure~\ref{fig:vis_results} presents the visualization results of two challenging scenarios from the OPV2V dataset. CoopDETR produces more accurate detection results compared to previous cooperative perception models. Specifically, the bounding boxes generated by CoopDETR exhibit better alignment with ground truth. This improvement shows our object-level fusion paradigm extracts more accurate representations of objects.

\section{CONCLUSIONS}

We propose CoopDETR, a novel query-based cooperative perception framework that leverages object queries to facilitate efficient communication and collaboration across multi-agent systems. By leveraging queries for object-level feature fusion, CoopDETR significantly reduces transmission costs while enhancing detection accuracy compared to other early fusin, late fusion, and regional-level feature fusion methods. Experiments on OPV2V and V2XSet datasets show state-of-the-art performance and robustness to pose errors, demonstrating superior performance in improving perception capability and communication efficiency for multi-agent cooperation.
Future research could focus on the integration of data from various sensing modalities and end-to-end framework~\cite{hu2023_uniad,yu2024_univ2x} that jointly optimizes perception, communication, and decision-making pipeline.

\section*{ACKNOWLEDGMENT}
This work is funded by the National Science and Technology Major Project (2022ZD0115502) and Lenovo Research.

\bibliographystyle{IEEEtran}
\balance
\bibliography{IEEEexample}

\end{document}